\documentclass[runningheads]{llncs}
\usepackage[T1]{fontenc}
\usepackage{graphicx,verbatim}
\usepackage{amsfonts}
\usepackage{amsmath}
\usepackage{bm}
\usepackage{multirow}
\usepackage{xcolor}
\usepackage{boldline}
\usepackage{marvosym}
\begin{document}
\title{MaskedCLIP: Bridging the Masked and CLIP Space for Semi-Supervised Medical Vision-Language Pre-training}
\titlerunning{MaskedCLIP for Semi-Supervised Medical Vision-Language Pre-training}

\author{Lei Zhu\inst{1\text{\Letter}} \and
Jun Zhou\inst{1} \and
Rick Siow Mong Goh\inst{1}\and
Yong Liu\inst{1}}
\authorrunning{Lei Zhu et al.}
\institute{Institute of High Performance Computing (IHPC), Agency for Science, Technology and Research (A*STAR), 1 Fusionopolis Way, \#16-16 Connexis, Singapore 138632, Republic of Singapore \\ \email{zhu\_lei@ihpc.a-star.edu.sg}}

\maketitle              
\begin{abstract}
Foundation models have recently gained tremendous popularity in medical image analysis. State-of-the-art methods leverage either paired image-text data via vision-language pre-training or unpaired image data via self-supervised pre-training to learn foundation models with generalizable image features to boost downstream task performance. However, learning foundation models exclusively on either paired or unpaired image data limits their ability to learn richer and more comprehensive image features. In this paper, we investigate a novel task termed semi-supervised vision-language pre-training, aiming to fully harness the potential of both paired and unpaired image data for foundation model learning. To this end, we propose \textbf{MaskedCLIP}, a synergistic masked image modeling and contrastive language-image pre-training framework for semi-supervised vision-language pre-training. The key challenge in combining paired and unpaired image data for learning a foundation model lies in the incompatible feature spaces derived from these two types of data. To address this issue, we propose to connect the masked feature space with the CLIP feature space with a bridge transformer. In this way, the more semantic specific CLIP features can benefit from the more general masked features for semantic feature extraction. We further propose a masked knowledge distillation loss to distill semantic knowledge of original image features in CLIP feature space back to the predicted masked image features in masked feature space. With this mutually interactive design, our framework effectively leverages both paired and unpaired image data to learn more generalizable image features for downstream tasks. Extensive experiments on retinal image analysis demonstrate the effectiveness and data efficiency of our method.


\end{abstract}

\section{Introduction}
Deep neural networks~\cite{litjens2017survey} have been a fundamental tool in medical image analysis, yet they often require a large amount of labeled training data to be effective and the models can sometimes be biased towards the semantic labels. Foundation models~\cite{bommasani2021opportunities} provide a promising approach to alleviate these issues via pre-training deep neural networks on diverse and large volume of medical image data to learn generalizable image features to boost downstream task performance. State-of-the-art (SoTA) pre-training methods can be generally categorized into vision-language pre-training methods~\cite{radford2021learning} and self-supervised pre-training methods~\cite{caron2020unsupervised,he2022masked,zhou2023foundation,oquab2023dinov2}. Contrastive Language-Image Pre-training (CLIP)\cite{radford2021learning} is a leading vision-language pre-training method in general image analysis, where it leverages large-scale paired image-text data and contrastively aligns image features with text features in a shared feature space. Building on CLIP, numerous studies in medical domain have trained foundation models for different image modalities, including Chest X-ray~\cite{wang2022medclip}, computed tomography (CT)~\cite{hamamci2024developing}, pathology~\cite{ikezogwo2024quilt}, among others. More recently, FLAIR~\cite{silva2025foundation} proposes to encode expert knowledge to the text branch of CLIP for retinal image analysis, which boosts CLIP model performance. Self-supervised pre-training methods propose various pretext tasks to learn foundation models with unpaired image data. In general image analysis, contrastive learning based methods, such as SimCLR~\cite{chen2020simple}, SwAV~\cite{caron2020unsupervised}, and MoCo-v3~\cite{chen2021empirical}, learn to maximize agreement between differently augmented samples in feature space. DINO~\cite{caron2021emerging} proposes self-distillation on multi-view images. Masked image modeling~\cite{he2022masked} pre-trains transformer to reconstruct masked image patches. iBot~\cite{zhou2021ibot} performs self-distillation on masked image patches with an online tokenizer. DINOv2~\cite{oquab2023dinov2} combines image-level~\cite{caron2021emerging} and patch-level~\cite{zhou2021ibot} self-distillation with a novel data curation pipeline, which achieves state-of-the-art performance for various downstream tasks. More recently, RETFound~\cite{zhou2023foundation} performs a systematic study to compare different self-supervised learning methods on retinal image analysis, where they found that generative based masked image modeling method~\cite{he2022masked} outperforms contrastive learning based ones~\cite{chen2020simple,caron2020unsupervised,caron2021emerging,chen2021empirical}.

\begin{figure}[t]
\centering
\includegraphics[width=0.55\textwidth]{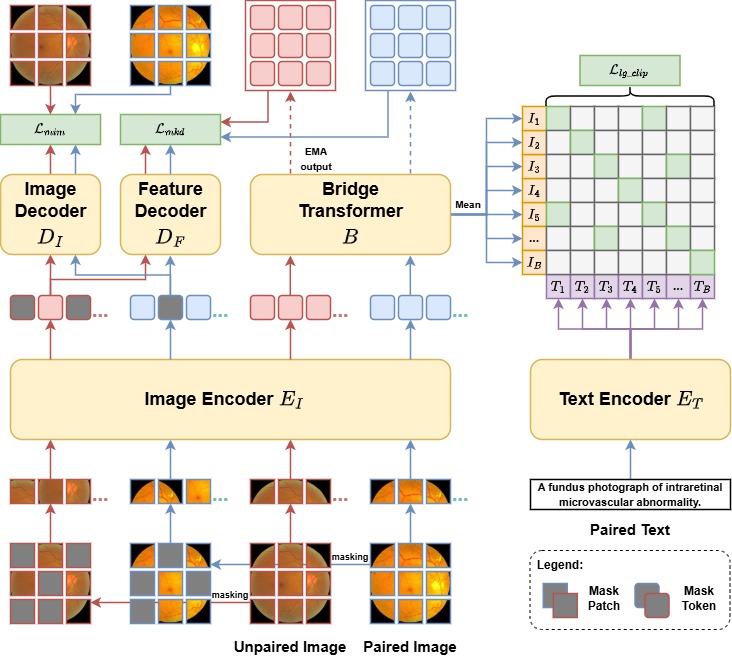}
\caption{Architecture and dataflow of our proposed MaskedCLIP framework. Our framework consists of five modules, namely an image encoder, a bridge transformer, a text encoder, an image decoder, and a feature decoder to process both image and text data for synergistic masked image modeling and contrastive language-image pre-training. We employ the bridge transformer to connect the masked and CLIP feature space to resolve the feature incompatibility issue and the feature decoder to predict masked image features for masked knowledge distillation.}
\label{fig1}
\end{figure}

While existing pre-training methods can train powerful foundation models to boost downstream task performance, they have focused exclusively on leveraging either paired or unpaired image data for learning foundation models, which limits their ability to learn richer and more comprehensive image features. In this paper, we investigate a novel task termed semi-supervised vision-language pre-training, aiming to fully harness the potential of both paired and unpaired image data for foundation model learning. To this end, we propose \textbf{MaskedCLIP}, a synergistic masked image modeling and contrastive language-image pre-training framework for semi-supervised vision-language pre-training. The key challenge in combining paired and unpaired image data for learning a foundation model lies in the incompatible feature spaces derived from these two types of data, where the CLIP feature space captures more semantic specific features, while the masked feature space retains more general image features. Thus, a naive approach to directly share the masked feature space with CLIP feature space suffers from the feature incompatibility issue and will result in poor performance. To address this issue, we propose to connect the masked feature space with the CLIP feature space with a bridge transformer. In this way, the more semantic specific CLIP features can benefit from the more general masked features for semantic feature extraction. We further propose a masked knowledge distillation loss to distill the semantic knowledge of original image features in CLIP feature space back to the predicted masked image features in the masked feature space. With this mutually interactive design, the masked features and CLIP features benefit from each other for feature representation learning, which enables our framework to learn more generalizable image features for downstream tasks.

In summary, we have made the following contributions in this paper: \textbf{(1).} We introduce a novel task termed semi-supervised vision-language pre-training for learning foundation models; \textbf{(2).} We propose MaskedCLIP, a principally designed framework for semi-supervised vision-language pre-training; \textbf{(3).} We conduct extensive experiments to evaluate the effectiveness of our method on retinal image analysis, where it significantly outperforms existing methods across seven downstream tasks and demonstrates exceptional label efficiency.

\section{Methodology}

In semi-supervised vision-language pre-training, we are give an assembly of paired image-text data with $N^p$ data points. We represent the paired image-text data in a triplet format to accommodate optional categorical labels as $\mathbb{D}^{p}=\{(x_{i}^p,t_{i}^p,y_{i}^p)\}^{N^p}_{i=1}$, where $t_i^p$ is the associated text description for $x_i^p$ and $y_i^p$ is the associated categorical label of $x_i^p$ if available; otherwise $y_i^p$ is defined as a unique identifier for the image-text pair. Additional, we are given an assembly of unpaired image data $\mathbb{D}^{u}=\{x_{i}^u\}^{N^u}_{i=1}$ with $N^u$ data points. The goal is effectively leverage both the paired and unpaired image data to train a foundation model. Fig.~\ref{fig1} presents an overview of our proposed MaskedCLIP framework.

\subsection{Bridging the Masked and CLIP Space}
Self-supervised pre-training and vision-language pre-training are two disparate learning paradigms that utilize either unpaired image data or paired image-text data for learning generalizable image feature representations. To effectively leverage both paired and unpaired image data for foundation model learning, one naive idea is to combine the two learning paradigms with
a shared image encoder to learn a common image feature space. However, we note that there exists a natural semantic hierarchical structure across the feature spaces learned from the two types of data, where the vision-language pre-trained image features are more semantic specific due to language supervision, while self-supervised pre-trained image features are more general due to lack of supervision signal. Thus, naively sharing the image encoder from the two learning paradigms suffers from the feature incompatibility issue and will result in poor performance. To address this, we propose to bridge the two feature spaces while preserving their semantic hierarchical structure so that the more semantic specific vision-language image features can benefit from the more general self-supervised image features for semantic feature extraction. In our framework, we propose synergistic masked image modeling~\cite{he2022masked} and contrastive language-image pre-training~\cite{radford2021learning} with a bridge transformer to bridge the masked and CLIP feature space.

Specifically, we utilize an image encoder $E_{I}$ together with an image decoder $D_{I}$ to construct the masked feature space for masked image modeling. We propose to combine paired image data together with the unpaired image data for the task to enhance data diversity. Following~\cite{he2022masked}, the image encoder takes only the visible image patches as input and the image decoder takes concatenated latent features from visible image patches and learnable mask tokens with positional embedding as input to reconstruct the masked image patches. The masked image modeling loss is defined as follow:
\begin{equation} \label{eqn:0} 
\mathcal{L}_{mim}= \frac{1}{|\mathcal{B}^p|+|\mathcal{B}^u|} \sum_{x\in \mathcal{B}^p \cup \mathcal{B}^u} \frac{1}{\mathcal{M}}\sum_{i\in \mathcal{M}}||x[i]-D_I([E_I(x_v);\bm{T_I}])[i]||^2,
\end{equation}
where $x_v$ denotes the visible image patches of $x$, $\bm{T_I}$ denotes the set of learnable mask tokens with positional embeddings for image pixel reconstruction, the operation $[\cdot;\cdot]$ concatenates two vectors into a single vector, $[\cdot]$ selects the indexed image patch from an image, $||\cdot||$ calculates the $l_2$-norm, $\mathcal{M}$ denotes the indexes of masked image patches, $\mathcal{B}^p$ and $\mathcal{B}^u$ denote batches of paired and unlabeled data sampled from $\mathbb{D}^p$ and $\mathbb{D}^u$ respectively.

Next, we introduce a bridge transformer $B$ to connect the masked and CLIP feature space and a text encoder $E_T$ to extract text features for contrastive language-image pre-training. Following~\cite{radford2021learning}, we employ a lightweight projection head at the end of both the bridge transformer and the text encoder to map the mean image and mean text features into a shared feature space. Vanilla contrastive loss~\cite{radford2021learning} does not consider the categorical labels of different images, which can lead to images with the same categorical labels being erroneously pushed apart from the paired text of another image. Inspired from~\cite{yang2022unified}, we perform label-guided contrastive learning, where we ensure image-text pairs are pulled together if they share the same categorical label. We utilize the paired image-text data for pre-training, where we employ the image-to-text contrastive loss to align matched images to a given text and text-to-image contrastive loss to align matched texts to a given image. The loss functions are defined as follow:
\begin{equation} \label{eqn:1} 
\mathcal{L}_{i2t}= \frac{1}{|\mathcal{B}^p|} \sum_{(x, t)\in \mathcal{B}^p} \frac{1}{|\mathcal{P}(x)|}\sum_{(x',t')\in \mathcal{P}(x)}log\frac{exp(\tau B(E_I(x))^TE_T(t'))}{\sum_{(x'', t'')\in \mathcal{B}^p}exp(\tau B(E_I(x))^TE_T(t''))},
\end{equation}
\begin{equation} \label{eqn:2} 
\mathcal{L}_{t2i}= \frac{1}{|\mathcal{B}^p|} \sum_{(x, t)\in \mathcal{B}^p} \frac{1}{|\mathcal{P}(x)|}\sum_{(x',t')\in \mathcal{P}(x)}log\frac{exp(\tau B(E_I(x'))^TE_T(t))}{\sum_{(x'', t'')\in \mathcal{B}^p}exp(\tau B(E_I(x''))^TE_T(t))},
\end{equation}
where $\tau$ is a learnable scaling parameter and $\mathcal{P}(x)=\{(x',t')|(x',t')\in \mathcal{B}^p, y'=y\}$ is the set of image-text pairs with same categorical label as $x$ within the batch.

The label-guided contrastive language-image pre-training loss is the combination of both image-to-text and text-to-image contrastive loss and is defined as follows:
\begin{equation} \label{eqn:3} 
\mathcal{L}_{lg\_clip}= \frac{1}{2} \mathcal{L}_{i2t} + \frac{1}{2} \mathcal{L}_{t2i}.
\end{equation}

\noindent \textbf{Discussion.} While label-guided contrastive learning requires categorical labels as input, it reduces to the vanilla contrastive loss function when categorical labels are not available. Additional, for paired image-label data, we can apply simple prompt to convert categorical labels into text descriptions~\cite{radford2021learning}. Thus, by incorporating label-guided contrastive learning into our framework, our framework works with both paired image-text and paired image-label data, which highlights the wide applicability of our approach in medical domain.

\subsection{Masked Knowledge Distillation}
We propose a masked knowledge distillation loss to further transfer semantic knowledge from original image features in CLIP feature space back to predicted masked image features in masked feature space. Such a loss function offers two key benefits: (1) It complements the pixel reconstruction loss in masked image modeling by guiding the model to extract semantic information from low-level image features for semantic feature reconstruction; (2) It enhances CLIP in global semantic information learning by extracting semantic information from local image patches. We leverage a feature decoder $D_F$ to reconstructs the masked image features and propose to extract robust image features from original images in CLIP space as targets with the momentum encoder $\hat{M}=EMA(B\circ E_I)$, where $EMA(\cdot)$ calculates the exponential moving average of the encoder. We combine both the paired and unpaired image data for masked knowledge distillation. We normalize the masked image patch features and the target image patch features and minimize the cosine distance for all image patches for knowledge distillation. The masked knowledge distillation loss is defined as follows:
\begin{equation} \label{eqn:4} 
\mathcal{L}_{mfd}= \frac{1}{|\mathcal{B}^p|+|\mathcal{B}^u|} \sum_{x\in \mathcal{B}^p \cup \mathcal{B}^u} \frac{1}{\mathcal{K}}\sum_{i\in \mathcal{K}}-\langle \frac{D_F([E_I(x_v);\bm{T_F}])[i]}{||D_F([E_I(x_v);\bm{T_F}])[i]||}, \frac{\hat{M}(x)[i]}{||\hat{M}(x)[i]||} \rangle,
\end{equation}
where $\bm{T_F}$ denotes the set of learnable mask tokens with positional embeddings for image feature reconstruction, $-\langle\cdot,\cdot\rangle$ calculates the cosine distance of two normalized vectors, and $\mathcal{K}$ denotes the indexes of all image patches. 

\noindent \textbf{Overall Objective.} The overall objective of our MaskedCLIP framework is defined as follows:
\begin{equation} \label{eqn:5} 
\mathcal{L}_{maskedclip}= \mathcal{L}_{min} + \lambda_{lg\_clip} \mathcal{L}_{lg\_clip} + \lambda_{mfd} \mathcal{L}_{mfd}.
\end{equation}
where $\lambda_{lg\_clip}$ and $\lambda_{mfd}$ are two balancing weights. We empirically tune these two hyper-parameters based on their magnitudes and set them to 0.01.

\section{Experimental Analysis}
\noindent \textbf{Pre-training Datasets.} We assemble a pre-training dataset with 15 public datasets and 10 private datasets for retinal image analysis. In total, the assembled dataset comprises 348,481 color fundus images.
The public datasets contain main retinal image analysis tasks in diabetic retinopathy grading~\cite{diabetic_retinopathy_kaggle,benitez2021dataset,liu2022deepdrid}, glaucoma detection~\cite{sivaswamy2014drishti,diaz2019cnns,orlando2020refuge,li2019attention,de2023airogs,huang2023grape,kumar2023chakṣu}, and some other disease diagnosis~\cite{budai2013robust,lin2020sustech,jin2022fives}. While most of the public datasets contain categorical labels, we also include two datasets that contain text descriptions: ODIR-5K~\cite{ODIR2019} and STARE~\cite{hoover2000locating}. We build the paired image-text data with three public datasets, namely ODIR-5K~\cite{ODIR2019}, AIROGS~\cite{de2023airogs}, and EYEPACS~\cite{diabetic_retinopathy_kaggle}, which contains 142,249 images in total. We follow~\cite{silva2025foundation} to encode expert knowledge to text descriptions when constructing image-text pair. We utilize the rest public datasets and all private datasets to build the unpaired image data.

\noindent \textbf{Downstream Tasks and Comparison Methods.} We evaluate the performance of our pre-trained foundation model on 7 public datasets across three retinal image analysis tasks: diabetic retinopathy grading (APTOS~\cite{aptos2019_kaggle}, IDRID~\cite{porwal2020idrid}, MESSIDOR-2~\cite{decenciere2014feedback}), glaucoma detection (GF~\cite{ahn2018deep}, ORIGA~\cite{zhang2010origa}), and multi-disease diagnosis (JSIEC~\cite{cen2021automatic}, Retina~\cite{cataractdataset_kaggle}). We employ two commonly-used classification metrics, namely the area under receiver operating curve (ROC) and the area under precision-recall curve (PRC) to quantitatively evaluate the downstream task performance. We compare our method with a baseline Random method without pre-training; SoTA self-supervised pre-training methods MAE~\cite{he2022masked} and DINOv2~\cite{oquab2023dinov2}, where both methods are pre-trained on all paired and unpaired image data in our assembled dataset; SoTA vision-language pre-training method CLIP~\cite{radford2021learning} which is pre-trained on the paired image data in our assembled dataset; a baseline supervised pre-training method ImageNet21K~\cite{deng2009imagenet}, which is supervised pre-trained on about 14M labeled general images; SoTA foundation models in retinal image analysis, namely RETFound~\cite{zhou2023foundation} pre-trained using MAE on about 0.9M color fundus images and FLAIR~\cite{silva2025foundation} pre-trained with encoded expert knowledge using CLIP on about 0.28M paired color fundus images.

\noindent \textbf{Implementation Details.} We implement the image encoder with ViT-large and the image decoder with ViT-small~\cite{dosovitskiy2020image}. All input image is resized to 224$\times$224. The patch size is set to 16$\times$16. The masked ratio is set to 0.75 following\cite{he2022masked}. The EMA parameter of the momentum encoder is set to 0.999. The bridge transformer and feature decoder are implemented using a vision transformer with 4 transformer blocks. The text encoder is implemented using BioClinicalBERT~\cite{alsentzer2019bioclinicalbert}. We train the model with AdamW optimizer for 200 epochs with a learning rate of 1.5e-4 and a warm-up period of 40 epochs. The model is trained on 4 NVIDIA A100 GPUs with a batch size of 720 (4 × 180 per GPU) for both paired and unpaired data. For downstream task fine-tuning, we initialize a ViT-large model with the pre-trained weights from our image encoder. We set the batch size to 16 and fine-tune the model with AdamW optimizer for 50 epochs with a learning rate of 5e-4 and a warm-up period of 10 epochs. 

\begin{table*}[t]
\caption{Comparison with SoTA methods and foundation models on different downstream tasks. The best results are in \textbf{bold}, and the second-best results are \underline{underlined}.} 
\label{table:sota_comparison}
\centering
\resizebox{0.9\linewidth}{!}{%
\begin{tabular}{c| c | c | c |  cc | cc |>{\centering\arraybackslash}p{0.95cm}c | cc| cc | cc | cc | cc}
\hlineB{3}
\multirow{2}{*}{Labeled} 
& \multirow{2}{*}{Method} 
& Paired & Unpaired & \multicolumn{2}{c|}{APTOS} & \multicolumn{2}{c|}{IDRID} & \multicolumn{2}{c|}{MESSIDOR-2} & \multicolumn{2}{c|}{GF} & \multicolumn{2}{c|}{ORIGA} 
& \multicolumn{2}{c|}{JSIEC} & \multicolumn{2}{c|}{Retina} & \multicolumn{2}{c}{Avg} \\
\cline{5-20}
&&Data Size & Data Size  & ROC & PRC & ROC & PRC & ROC & PRC & ROC & PRC & ROC & PRC & ROC & PRC & ROC & PRC & ROC & PRC  \\
\hline
\multirow{8}{*}{10\%} 
&Random&0&0&73.1&34.6&53.0&28.9&68.1&29.1&81.3&64.1&52.2&54.0&64.8&10.5&59.7&38.7&64.6&37.1 \\
\cline{2-20}
&CLIP~\cite{radford2021learning}&0.14M&0&87.9&51.6&61.0&33.2&78.7&37.1&87.9&73.1&58.8&57.2&79.1&26.8&62.5&41.6&73.7&45.8 \\
&DINOv2~\cite{oquab2023dinov2}&0&0.34M&88.1&53.8&67.1&35.4&74.6&38.4&\textbf{90.5}&\textbf{76.7}&53.4&50.7&83.8&28.3&60.4&35.9&74.0&45.6 \\
&MAE~\cite{he2022masked}&0&0.34M&\underline{91.9}&58.8&\underline{72.7}&\underline{40.7}&\underline{79.2}&\underline{44.3}&86.8&73.5&53.2&54.2&85.6&38.2&67.7&\underline{48.7}&76.7&51.2 \\
\cline{2-20}
&ImageNet21K~\cite{deng2009imagenet}&14M&0&89.2&54.5&71.9&39.2&77.0&42.6&87.6&71.2&62.5&60.3&\underline{85.6}&\textbf{42.0}&63.9&40.7&76.8&50.1 \\
&FLAIR~\cite{silva2025foundation}&0.28M&0&90.2&54.3&63.9&33.6&76.2&41.7&84.8&67.1&59.3&57.6&83.1&31.6&68.3&47.6&75.1&47.7 \\
&RETFound~\cite{zhou2023foundation}&0&0.90M&91.4&\underline{61.9}&66.8&38.1&78.1&41.8&\underline{88.6}&\underline{74.1}&\underline{65.4}&\underline{62.9}&\textbf{87.1}&\underline{41.5}&\underline{68.5}&45.9&\underline{78.0}&\underline{52.3} \\
\cline{2-20}
&MaskedCLIP&0.14M&0.20M&\textbf{93.7}&\textbf{66.3}&\textbf{78.4}&\textbf{52.3}&\textbf{84.3}&\textbf{56.6}&88.0&71.9&\textbf{72.5}&\textbf{66.8}&85.2&35.8&\textbf{72.4}&\textbf{54.7}&\textbf{82.1}&\textbf{57.8} \\
\hlineB{4}
\multirow{8}{*}{100\%} 
&Random&0&0&83.7&45.8&56.6&30.5&69.9&30.4&87.1&72.0&52.4&52.9&81.9&27.9&61.8&40.8&70.5&42.9\\
\cline{2-20}
&CLIP~\cite{radford2021learning}&0.14M&0&92.0&64.6&75.9&44.8&83.9&52.8&92.4&82.5&61.6&58.7&97.6&76.5&82.8&66.5&83.7&63.8\\
&DINOv2~\cite{oquab2023dinov2}&0&0.34M&94.2&70.7&77.1&46.8&85.2&59.5&\textbf{95.2}&\textbf{88.5}&64.0&58.1&99.3&89.6&81.9&66.3&85.3&68.5\\
&MAE~\cite{he2022masked}&0&0.34M&94.0&71.7&80.9&48.7&87.7&59.7&91.9&81.7&\underline{72.1}&\underline{65.4}&99.3&89.7&\underline{85.4}&\underline{69.0}&\underline{87.3}&69.4\\
\cline{2-20}
&ImageNet21K~\cite{deng2009imagenet}&14M&0&94.3&70.3&78.0&47.6&86.3&60.9&91.5&80.7&64.8&60.4&\textbf{99.7}&\textbf{93.4}&80.5&60.6&85.0&67.7\\
&FLAIR~\cite{silva2025foundation}&0.28M&0&93.4&68.1&75.8&47.5&86.2&57.1&90.6&78.9&66.0&60.9&99.1&84.6&81.5&59.2&84.7&65.2\\
&RETFound~\cite{zhou2023foundation}&0&0.90M&\textbf{95.0}&\textbf{74.4}&\underline{83.0}&\underline{51.3}&\underline{88.1}&\underline{65.0}&\underline{94.1}&\underline{85.2}&68.1&62.6&99.0&89.8&83.4&68.4&87.2&\underline{71.0}\\
\cline{2-20}
&MaskedCLIP&0.14M&0.20M&\underline{94.8}&\underline{73.4}&\textbf{83.1}&\textbf{56.3}&\textbf{88.8}&\textbf{68.5}&93.4&85.0&\textbf{72.8}&\textbf{66.2}&\underline{99.5}&\underline{91.9}&\textbf{90.4}&\textbf{79.5}&\textbf{89.0}&\textbf{74.4}\\
\hlineB{3}
\end{tabular}}
\end{table*}

\begin{table*}[t]
\caption{Ablation study on different downstream tasks with 10\% training data. The best results are in \textbf{bold}, and the second-best results are \underline{underlined}.} 
\label{table:ablation}
\centering
\resizebox{0.9\linewidth}{!}{%
\begin{tabular}{ c | c  >{\centering\arraybackslash}p{1cm}|  cc | cc | >{\centering\arraybackslash}p{0.95cm}c | cc| cc | cc | cc | cc}
\hlineB{3}
\multirow{2}{*}{Method} 
& Bridge &\multirow{2}{*}{$\mathcal{L}_{mfd}$} & \multicolumn{2}{c|}{APTOS} & \multicolumn{2}{c|}{IDRID} & \multicolumn{2}{c|}{MESSIDOR-2} & \multicolumn{2}{c|}{GF} & \multicolumn{2}{c|}{ORIGA} 
& \multicolumn{2}{c|}{JSIEC} & \multicolumn{2}{c|}{Retina} & \multicolumn{2}{c}{Avg} \\
\cline{4-19}
&Transformer& & ROC & PRC & ROC & PRC & ROC & PRC & ROC & PRC & ROC & PRC & ROC & PRC & ROC & PRC & ROC & PRC  \\
\hline
MAE+CLIP&&&\underline{91.5}&59.3&\underline{72.6}&\underline{39.2}&82.1&43.2&\underline{87.8}&\underline{71.8}&66.7&60.9&81.3&26.9&65.7&42.6&78.2&49.2\\
+Bridge Transformer&\checkmark&&\underline{91.5}&\underline{61.1}&71.0&37.1&\underline{82.9}&\underline{44.0}&85.0&68.5&\underline{72.1}&\underline{65.2}&\underline{82.2}&\underline{30.4}&\underline{68.5}&\underline{46.0}&\underline{79.0}&\underline{50.3}\\
MaskedCLIP&\checkmark&\checkmark&\textbf{93.7}&\textbf{66.3}&\textbf{78.4}&\textbf{52.3}&\textbf{84.3}&\textbf{56.6}&\textbf{88.0}&\textbf{71.9}&\textbf{72.5}&\textbf{66.8}&\textbf{85.2}&\textbf{35.8}&\textbf{72.4}&\textbf{54.7}&\textbf{82.1}&\textbf{57.8}\\
\hlineB{3}
\end{tabular}}
\end{table*}

\begin{figure}[t]
\centering
\includegraphics[width=0.85\textwidth]{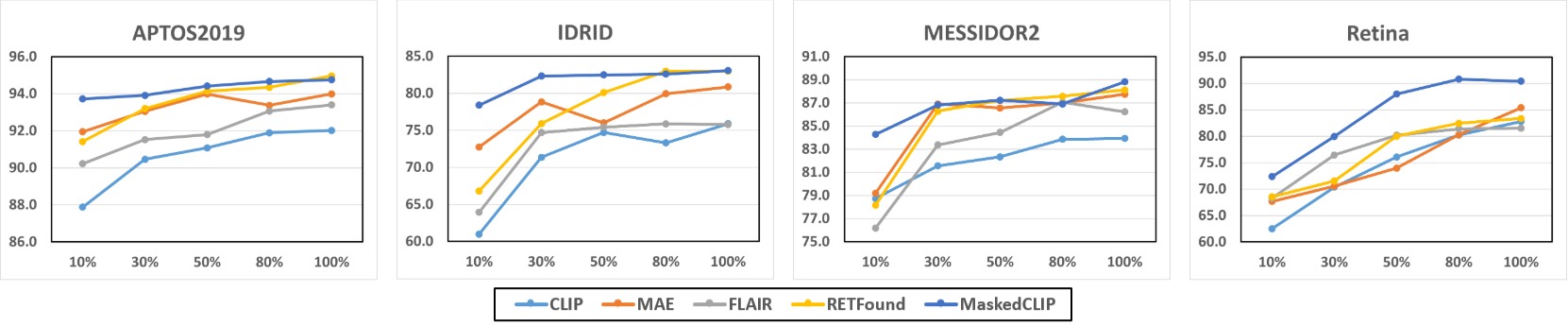}
\caption{Label efficiency analysis on exemplary downstream tasks. The X axis shows the training data proportion and the Y axis shows the ROC score.}
\label{fig2}
\end{figure}

\noindent \textbf{Comparison with SoTA Methods.} In Table~\ref{table:sota_comparison}, we compare our method with SoTA methods and foundation models under two learning scenarios, namely a label scarce setting with 10\% training data and a label abundant setting with entire training data for fine-tuning across 7 downstream tasks. In both scenarios, our method consistently outperforms or matches existing methods and foundation models with significantly better average ROC and PRC scores. Specifically, our method outperforms SoTA self-supervised pre-training methods DINOv2 and MAE and vision-language pre-training method CLIP. Either DINOv2 and MAE or CLIP can only leverage unpaired or paired image data for pre-training, which limits their ability to learn richer and more comprehensive image features. In contrast, our method effectively integrates both paired and unpaired image data for foundation model learning, leading to significantly better performance. \textbf{The experiment results highlight that rather than focusing exclusively on either paired or unpaired image data, a unified approach that leverages all available image data is key to develop more powerful foundation models.} Furthermore, our method consistently outperforms or matches SoTA foundation models FLAIR and RETFound despite they are pre-trained with much larger paired or unpaired image datasets. The experiment results demonstrate that our method is more data efficient for foundation model learning than existing methods. We attribute this advantage to the integration of both paired and unpaired image data for pre-training and the novel mutually interactive design of our framework where the masked feature space is bridged to support CLIP feature space for semantic feature extraction and the CLIP feature space guides masked feature space for semantic feature learning through masked knowledge distillation. This mutually interactive design maximizes the utilization of both types of data. Finally, our method achieves the most improvement when compared to the second best in label scare setting, which highlights the label efficiency of our method for downstream tasks.

\noindent \textbf{Ablation Study.} In Table~\ref{table:ablation}, we present an ablation study on different components of our method. As observed, directly combining MAE and CLIP by sharing their image encoder results in suboptimal performance. Introducing a bridge transformer to address the feature incompatibility issue between the two feature spaces significantly improves performance across multiple datasets. Finally, further incorporating masked knowledge distillation effectively enables the mutual interaction between the masked and CLIP feature spaces, which leads to the best performance across all downstream tasks.

\noindent \textbf{More Label Efficiency Analysis.} In Fig.~\ref{fig2}, we present a more detailed label efficiency analysis of our method against SoTA methods on APTOS2019, IDRID, MESSIDOR2, and Retina datasets. As shown, our method consistently outperforms or matches existing methods and foundation models in all training data proportions, highlighting its wide applicability in diverse learning scenarios.

\section{Conclusion}
In this paper, we introduce a novel task termed semi-supervised vision-language pre-training and propose MaskedCLIP, a principally designed framework to fully harness the potential of both paired and unpaired image data for foundation model learning. While existing studies have mostly focused on leveraging only paired or unpaired image data for learning foundation models, we advocate for a unified approach that integrates all available image data, either paired or unpaired to develop more powerful foundation models in medical domain. Our approach demonstrates promising results and we hope our work can inspire future studies to further explore this direction.

\noindent \textbf{Acknowledgement}
This work was supported by the Agency for Science, Technology, and Research (A*STAR) through its IEO Decentralised GAP Under Project I24D1AG085.

\newpage
\pagebreak
%
%
%
%
\bibliographystyle{splncs04}
\bibliography{reference}

\begin{thebibliography}{10}
\providecommand{\url}[1]{\texttt{#1}}
\providecommand{\urlprefix}{URL }
\providecommand{\doi}[1]{https://doi.org/#1}

\bibitem{ahn2018deep}
Ahn, J.M., Kim, S., Ahn, K.S., Cho, S.H., Lee, K.B., Kim, U.S.: A deep learning model for the detection of both advanced and early glaucoma using fundus photography. PloS one  \textbf{13}(11),  e0207982 (2018)

\bibitem{benitez2021dataset}
Ben{\'\i}tez, V.E.C., Matto, I.C., Rom{\'a}n, J.C.M., Noguera, J.L.V., Garc{\'\i}a-Torres, M., Ayala, J., Pinto-Roa, D.P., Gardel-Sotomayor, P.E., Facon, J., Grillo, S.A.: Dataset from fundus images for the study of diabetic retinopathy. Data in brief  (2021)

\bibitem{bommasani2021opportunities}
Bommasani, R., Hudson, D.A., Adeli, E., Altman, R., Arora, S., von Arx, S., Bernstein, M.S., Bohg, J., Bosselut, A., Brunskill, E., et~al.: On the opportunities and risks of foundation models. arXiv preprint arXiv:2108.07258  (2021)

\bibitem{budai2013robust}
Budai, A., Bock, R., Maier, A., Hornegger, J., Michelson, G.: Robust vessel segmentation in fundus images. International journal of biomedical imaging  (2013)

\bibitem{caron2020unsupervised}
Caron, M., Misra, I., Mairal, J., Goyal, P., Bojanowski, P., Joulin, A.: Unsupervised learning of visual features by contrasting cluster assignments. Advances in neural information processing systems  \textbf{33},  9912--9924 (2020)

\bibitem{caron2021emerging}
Caron, M., Touvron, H., Misra, I., J{\'e}gou, H., Mairal, J., Bojanowski, P., Joulin, A.: Emerging properties in self-supervised vision transformers. In: CVPR (2021)

\bibitem{cen2021automatic}
Cen, L.P., Ji, J., Lin, J.W., Ju, S.T., Lin, H.J., Li, T.P., Wang, Y., Yang, J.F., Liu, Y.F., Tan, S., et~al.: Automatic detection of 39 fundus diseases and conditions in retinal photographs using deep neural networks. Nature communications  (2021)

\bibitem{chen2020simple}
Chen, T., Kornblith, S., Norouzi, M., Hinton, G.: A simple framework for contrastive learning of visual representations. In: ICML (2020)

\bibitem{chen2021empirical}
Chen, X., Xie, S., He, K.: An empirical study of training self-supervised vision transformers. In: CVPR (2021)

\bibitem{de2023airogs}
De~Vente, C., Vermeer, K.A., Jaccard, et~al.: Airogs: Artificial intelligence for robust glaucoma screening challenge. IEEE transactions on medical imaging  (2023)

\bibitem{decenciere2014feedback}
Decenci{\`e}re, E., Zhang, X., Cazuguel, G., Lay, B., Cochener, B., Trone, C., Gain, P., Ord{\'o}{\~n}ez-Varela, J.R., Massin, P., E.A., et~al.: Feedback on a publicly distributed image database: the messidor database. Image Analysis \& Stereology  (2014)

\bibitem{deng2009imagenet}
Deng, J., Dong, W., Socher, R., Li, L.J., Li, K., Fei-Fei, L.: Imagenet: A large-scale hierarchical image database. In: CVPR (2009)

\bibitem{diaz2019cnns}
Diaz-Pinto, A., Morales, S., Naranjo, V., K{\"o}hler, T., Mossi, J.M., Navea, A.: Cnns for automatic glaucoma assessment using fundus images: an extensive validation. Biomedical engineering online  \textbf{18},  1--19 (2019)

\bibitem{dosovitskiy2020image}
Dosovitskiy, A.: An image is worth 16x16 words: Transformers for image recognition at scale. arXiv preprint arXiv:2010.11929  (2020)

\bibitem{alsentzer2019bioclinicalbert}
{Emily Alsentzer}:  (2019), \url{https://huggingface.co/emilyalsentzer/Bio\_ClinicalBERT}

\bibitem{hamamci2024developing}
Hamamci, I.E., Er, S., Almas, F., Simsek, A.G., Esirgun, S.N., Dogan, I., Dasdelen, M.F., Durugol, O.F., Wittmann, B., Amiranashvili, T., et~al.: Developing generalist foundation models from a multimodal dataset for 3d computed tomography  (2024)

\bibitem{he2022masked}
He, K., Chen, X., Xie, S., Li, Y., Doll{\'a}r, P., Girshick, R.: Masked autoencoders are scalable vision learners. In: CVPR (2022)

\bibitem{hoover2000locating}
Hoover, A., Kouznetsova, V., Goldbaum, M.: Locating blood vessels in retinal images by piecewise threshold probing of a matched filter response. TMI  (2000)

\bibitem{huang2023grape}
Huang, X., Kong, X., et~al.: Grape: A multi-modal dataset of longitudinal follow-up visual field and fundus images for glaucoma management. Scientific Data  (2023)

\bibitem{ikezogwo2024quilt}
Ikezogwo, W., Seyfioglu, S., et~al.: Quilt-1m: One million image-text pairs for histopathology. Neurips  (2024)

\bibitem{jin2022fives}
Jin, K., Huang, X., Zhou, J., Li, Y., Yan, Y., Sun, Y., Zhang, Q., Wang, Y., Ye, J.: Fives: A fundus image dataset for artificial intelligence based vessel segmentation. Scientific data  \textbf{9}(1), ~475 (2022)

\bibitem{diabetic_retinopathy_kaggle}
{Kaggle}:  (2015), \url{https://www.kaggle.com/c/diabetic-retinopathy-detection}

\bibitem{aptos2019_kaggle}
{Kaggle}:  (2019), \url{https://www.kaggle.com/c/aptos2019-blindness-detection}

\bibitem{cataractdataset_kaggle}
{Kaggle}:  (2022), \url{https://www.kaggle.com/datasets/jr2ngb/cataractdataset}

\bibitem{kumar2023chakṣu}
Kumar, J.H., Seelamantula, C.S., Gagan, J., Kamath, Y.S., Kuzhuppilly, N.I., Vivekanand, U., Gupta, P., Patil, S.: Ch{\'a}kṣu: A glaucoma specific fundus image database. Scientific data  \textbf{10}(1), ~70 (2023)

\bibitem{li2019attention}
Li, L., Xu, M., Wang, X., Jiang, L., Liu, H.: Attention based glaucoma detection: A large-scale database and cnn model. In: Proceedings of the IEEE/CVF conference on computer vision and pattern recognition. pp. 10571--10580 (2019)

\bibitem{lin2020sustech}
Lin, L., Li, M., Huang, Y., Cheng, P., Xia, H., Wang, K., Yuan, J., Tang, X.: The sustech-sysu dataset for automated exudate detection and diabetic retinopathy grading. Scientific Data  \textbf{7}(1), ~409 (2020)

\bibitem{litjens2017survey}
Litjens, G., Kooi, T., Bejnordi, B.E., et~al.: A survey on deep learning in medical image analysis. Medical image analysis  (2017)

\bibitem{liu2022deepdrid}
Liu, R., Wang, X., Wu, Q., Dai, L., Fang, X., et~al.: Deepdrid: Diabetic retinopathy—grading and image quality estimation challenge. Patterns  (2022)

\bibitem{ODIR2019}
{ODIR 2019 Grand Challenge}:  (2019), \url{https://odir2019.grand-challenge.org/}

\bibitem{oquab2023dinov2}
Oquab, M., Darcet, T., et~al.: Dinov2: Learning robust visual features without supervision. arXiv preprint arXiv:2304.07193  (2023)

\bibitem{orlando2020refuge}
Orlando, J.I., Fu, H., Breda, J.B., Van~Keer, K., et~al.: Refuge challenge: A unified framework for evaluating automated methods for glaucoma assessment from fundus photographs. Medical image analysis  (2020)

\bibitem{porwal2020idrid}
Porwal, P., Pachade, S., Kokare, M., Deshmukh, G., Son, J., Bae, W., Liu, L., Wang, J., Liu, X., Gao, L., et~al.: Idrid: Diabetic retinopathy--segmentation and grading challenge. Medical image analysis  \textbf{59},  101561 (2020)

\bibitem{radford2021learning}
Radford, A., Kim, J.W., Hallacy, C., Ramesh, A., Goh, G., Agarwal, S., Sastry, G., Askell, A., Mishkin, P., Clark, J., et~al.: Learning transferable visual models from natural language supervision. In: ICML (2021)

\bibitem{silva2025foundation}
Silva-Rodriguez, J., Chakor, H., Kobbi, R., Dolz, J., Ayed, I.B.: A foundation language-image model of the retina (flair): Encoding expert knowledge in text supervision. Medical Image Analysis  \textbf{99},  103357 (2025)

\bibitem{sivaswamy2014drishti}
Sivaswamy, J., Krishnadas, S., Joshi, G.D., Jain, M., Tabish, A.U.S.: Drishti-gs: Retinal image dataset for optic nerve head (onh) segmentation. In: ISBI (2014)

\bibitem{wang2022medclip}
Wang, Z., Wu, Z., Agarwal, D., Sun, J.: Medclip: Contrastive learning from unpaired medical images and text. arXiv preprint arXiv:2210.10163  (2022)

\bibitem{yang2022unified}
Yang, J., Li, C., Zhang, P., Xiao, B., Liu, C., Yuan, L., Gao, J.: Unified contrastive learning in image-text-label space. In: Proceedings of the IEEE/CVF Conference on Computer Vision and Pattern Recognition. pp. 19163--19173 (2022)

\bibitem{zhang2010origa}
Zhang, Z., Yin, F.S., et~al.: Origa-light: An online retinal fundus image database for glaucoma analysis and research. In: EMBC (2010)

\bibitem{zhou2021ibot}
Zhou, J., Wei, C., Wang, H., Shen, W., Xie, C., Yuille, A., Kong, T.: ibot: Image bert pre-training with online tokenizer. arXiv preprint arXiv:2111.07832  (2021)

\bibitem{zhou2023foundation}
Zhou, Y., Chia, M.A., Wagner, S.K., Ayhan, M.S., Williamson, D.J., Struyven, R.R., Liu, T., Xu, M., Lozano, M.G., Woodward-Court, P., et~al.: A foundation model for generalizable disease detection from retinal images. Nature  (2023)

\end{thebibliography}
\end{document}